\documentclass[10pt,twocolumn,letterpaper]{article}

\usepackage[pagenumbers]{cvpr}   

\usepackage{amsmath}
\usepackage{amsfonts}
\usepackage{multirow}
\usepackage{makecell}
\usepackage{booktabs}
\usepackage{graphicx}
\usepackage{animate}
\usepackage{pifont}
\usepackage{enumitem}
\usepackage[most]{tcolorbox}
\PassOptionsToPackage{table}{xcolor}
\usepackage{xcolor}
\usepackage{microtype}
\usepackage{url}            
\usepackage{nicefrac}       
\usepackage{colortbl}    
\usepackage{cuted}       
\usepackage{capt-of}     
\usepackage{float}       
\definecolor{highlightcolor}{HTML}{D0E8FF}
\newcommand{\cmark}{\textcolor{green!60!black}{\ding{51}}}
\newcommand{\xmark}{\textcolor{red!70!black}{\ding{55}}}

\definecolor{cvprblue}{rgb}{0.21,0.49,0.74}
\usepackage[pagebackref,breaklinks,colorlinks,allcolors=cvprblue]{hyperref}

\title{PAI-Actor: Cinematic Multi-Character Replacement in Dynamic Scenes}

\author{
  Bangxun Tang\textsuperscript{1,2,*} \quad
  Heyuan Gao\textsuperscript{1,3,*} \quad
  Yiren Song\textsuperscript{1,4,*,\ensuremath{\ddagger}} \quad
  Guian Fang\textsuperscript{1,4} \\
  Zijian He\textsuperscript{1} \quad
  Jie Yang\textsuperscript{1} \quad
  Mike Zheng Shou\textsuperscript{4,\ensuremath{\dagger}} \\
  \\
  \textsuperscript{1}Utopai Studios \quad
  \textsuperscript{2}University of California, Irvine \\
  \textsuperscript{3}Nanyang Technological University \quad
  \textsuperscript{4}Show Lab, National University of Singapore
}

\begin{document}
\maketitle

\begingroup
\renewcommand{\thefootnote}{}
\footnotetext{
\textsuperscript{*}Equal contribution.
\textsuperscript{$\ddagger$}Project leader.
\textsuperscript{$\dagger$}Corresponding author.
}
\endgroup

\begin{strip}
\centering
\animategraphics[width=\linewidth]{5}{fig/teaser/teaser_}{00}{09}%
\captionof{figure}{Given a source video and a reference image per character, PAI-Actor replaces characters in cinematic dynamic scenes while preserving background, camera motion, pose-accurate reenactment at 1080P resolution. Readers can click and play the video clips in this figure using {\color{red}\textbf{Adobe Acrobat}}.}
\label{fig:teaser}
\end{strip}

\begin{abstract}
We present PAI-Actor, a cinematic multi-character animation framework for character replacement in dynamic movie scenes. Unlike conventional animation systems that mainly drive a single static image or a single subject, our goal is to replace and animate multiple characters within real video clips while preserving the original scene dynamics, camera motion, and background content. This setting is particularly challenging because the generated characters must remain consistent with the source performance in motion and interaction, while also matching the surrounding background in lighting, shadow, composition, and overall cinematic appearance. To address this, we formulate multi-character animation as a structure-guided human recovery problem and build a movie-driven training pipeline from high-quality film data. Furthermore, to support practical cinematic production, we introduce a bidirectional-to-autoregressive distillation framework: we first train a bidirectional diffusion transformer for high-quality short-clip generation at 1080P resolution, and then distill it into an autoregressive video-to-video model for efficient inference and longer video generation. Experiments show that PAI-Actor enables high-fidelity multi-character animation with strong scene consistency, cinematic visual quality, and efficient long-form generation. Project page: \url{https://github.com/showlab/PAI-Actor}.
\end{abstract}

\section{Introduction}
\label{sec:intro}

Human animation has achieved impressive progress in recent years, yet cinematic multi-character animation remains a highly challenging problem. In real film production, character replacement is a common and practical demand: for example, motion-capture actors may perform in front of a green screen and later be transformed into cinematic characters or virtual figures. The goal is therefore not merely to animate a single static portrait or reenact one isolated subject. Instead, one often needs to replace and animate one or more characters inside an existing video clip, while preserving the original scene structure, camera movement, actor interactions, and background dynamics. This setting introduces a much more demanding form of controllable generation, where multiple characters must be re-rendered in a way that remains faithful to the source performance and visually coherent with the surrounding movie scene.

This task is exceptionally challenging. Generated characters must preserve the original motion, pose evolution, and multi-person interactions, while strictly maintaining scene consistency---matching illumination, shadows, and composition without altering the background. Furthermore, practical cinematic applications demand high-resolution, temporally stable, long-form videos rather than short clips. As summarized in Table~\ref{tab:feature}, existing pipelines struggle to jointly satisfy high-resolution multi-character generation, long-form scalability, and background preservation. Most notably, they uniformly fail to handle complex occlusions, a critical requirement for realistic dynamic scenes.

\begin{table*}[!t]
\centering
\caption{Feature comparison between PAI-Actor and existing character-replacement approaches.}
\label{tab:feature}
\footnotesize
\setlength{\tabcolsep}{4pt}
\renewcommand{\arraystretch}{1.0}
\resizebox{0.9\linewidth}{!}{
\begin{tabular}{lcccccccc>{\columncolor{highlightcolor}}c}
\toprule
\textbf{Feature}
& \textbf{Wan-Animate}
& \textbf{MultiAnimate}
& \textbf{SCAIL}
& \textbf{DOMO}
& \textbf{Viggle}
& \textbf{Kling O1}
& \textbf{MoCha}
& \textbf{SCAIL-2}
& \textbf{Ours} \\
\midrule
1080P                  & \xmark & \xmark & \xmark & \xmark & \cmark & \cmark & \cmark & \xmark & \cmark \\
Multi-Character            & \xmark & \cmark & \cmark & \cmark & \cmark & \cmark & \xmark & \cmark & \cmark \\
Long-Video Output ($>$10\,s)      & \xmark & \xmark & \xmark & \cmark & \cmark & \cmark & \xmark & \cmark & \cmark \\
Pose-Driven Control    & \cmark & \cmark & \cmark & \cmark & \xmark & \cmark & \xmark & \cmark & \cmark \\
Background Preservation& \cmark & \xmark & \xmark & \cmark & \cmark & \cmark & \cmark & \cmark & \cmark \\
Occlusion Handling     & \xmark & \xmark & \xmark & \xmark & \xmark & \xmark & \xmark & \xmark & \cmark \\
\bottomrule
\end{tabular}}
\end{table*}

In this work, we present PAI-Actor, a movie-driven framework for cinematic multi-character animation. Our key idea is to cast the task as a structure-guided human recovery problem. Given a movie clip, we segment and mask out the original human performers, extract structured motion signals through body pose association and face association, and render the corresponding skeletons on the masked regions as input. The model is then trained to recover the complete human performers in the original video space. This formulation allows the model to learn how humans should be re-rendered inside complex movie scenes, while preserving scene layout, background continuity, and interaction dynamics. More importantly, it naturally supports downstream multi-character replacement, where the original performers can be replaced by new target characters within the same dynamic scene.

To make the framework practical for cinematic production, we further introduce a bidirectional-to-autoregressive distillation strategy, which is a central component of our method. We first train a bidirectional video diffusion transformer as a teacher model to obtain high-quality generation results for short clips at 1080P resolution, typically around four seconds. While this bidirectional model provides strong visual fidelity and temporal modeling capacity, it is still expensive for long-video generation. We therefore distill it into an autoregressive video-to-video generator that supports much longer outputs. In this way, the distilled model inherits the quality of the teacher while significantly improving generation efficiency and temporal scalability. This design makes PAI-Actor suitable not only for high-quality short-clip synthesis, but also for practical long-form cinematic animation.

Our contributions are summarized as follows:
\begin{itemize}
    \item We propose PAI-Actor, the first movie-driven framework for cinematic multi-character replacement and animation that enables high-resolution, long-video character replacement in dynamic scenes with strong motion fidelity and scene consistency.
    \item We formulate multi-character animation as a structure-guided human recovery problem, and further introduce a bidirectional-to-autoregressive distillation framework that enables efficient inference and scalable long-form video generation.
    \item Our framework supports cinematic-quality multi-character animation with high resolution and scalable long-video generation, providing a practical solution for real-world character replacement and virtual production.
\end{itemize}


\section{Related Work}

\paragraph{Video Diffusion Models.}
Diffusion models~\cite{ho2020denoising,song2021score} have become a dominant paradigm for visual generation, with latent diffusion reducing the computational burden of high-resolution synthesis~\cite{rombach2022high} and diffusion transformers further improving scalability~\cite{peebles2023scalable}. Extending diffusion models from images to videos requires modeling both spatial fidelity and temporal coherence. Early video diffusion systems explore pixel-space or cascaded generation for text-to-video synthesis~\cite{ho2022video,ho2022imagen,singer2022make}, while subsequent latent-space approaches improve efficiency and resolution through temporal modules, hierarchical generation, or cascaded video latent diffusion~\cite{zhou2022magicvideo,he2022latent,blattmann2023align,wang2023lavie,chen2023videocrafter1,blattmann2023stable}. Another line adapts powerful image diffusion priors to video through one-shot tuning or zero-shot temporal attention mechanisms~\cite{wu2023tune,khachatryan2023text2videozero}. More recent models introduce space-time architectures, latent diffusion transformers, window attention, autoregressive token modeling, or expert transformers to enhance motion quality and long-range coherence~\cite{bartal2024lumiere,ma2024latte,gupta2024walt,kondratyuk2024videopoet,yang2024cogvideox}. Related directions investigate controllable diffusion transformers for editing and multi-domain sequence generation~\cite{feng2025dit4edit,song2025makeanything,zhang2025easycontrol}, as well as video generation across egocentric and exocentric views~\cite{song2025worldwander}. Despite rapid progress, most video diffusion models are designed for open-ended generation or coarse image-to-video synthesis. They do not directly solve cinematic character replacement, where multiple actors must be re-rendered with pose-accurate motion, stable identities, preserved camera motion, and unchanged dynamic backgrounds.

\paragraph{Animation and Motion-Driven Generation.}
Animation and motion-driven generation aim to synthesize videos conditioned on a driving signal such as keypoints, skeletons, dense motion, 3D parameters, audio, or a reference video. Classical image animation methods transfer motion from a driving video to a source image using learned keypoints, dense motion fields, articulated regions, or geometric warping~\cite{siarohin2019firstorder,chan2019everybody,siarohin2021motion,zhao2022thinplate}, while portrait animation methods use semantic or 3D facial parameters to obtain more disentangled control~\cite{ren2021pirenderer,zhang2023sadtalker}. With diffusion priors, recent human animation methods achieve stronger realism and temporal stability by combining reference appearance encoders, pose guiders, motion modules, disentangled foreground/background control, or 3D body guidance~\cite{wang2024disco,ma2024follow,karras2023dreampose,guo2024animatediff,xu2024magicanimate,hu2024animateanyone,zhu2024champ,zhang2024mimicmotion}. Recent work further studies regional motion control, motion transfer, selective subject representations, and visual-relation transfer~\cite{ma2025followyourclick,ma2025followyourmotion,zhang2024ssr,gong2025relationadapter}, while cross-embodiment video generation transfers human performance to robotic or humanoid embodiments~\cite{song2025mitty,yang2025x,song2026omnihumanoid}. Motion customization further shows that motion and appearance can be partially decoupled in text-to-video diffusion models~\cite{zhao2024motiondirector}. However, most existing animation systems focus on single-person reenactment, talking-head animation, fashion videos, or human dance. They typically animate one reference identity against a simple or weakly changing background, and therefore struggle with multi-actor interactions, occlusions, identity separation, and scene-level consistency in dynamic movie clips.

\paragraph{Video Editing.}
Video editing with generative models seeks to modify an existing video while preserving its layout, motion, and temporal coherence. Early text-driven editing methods demonstrate that local visual changes can be represented through layered edits, attention control, instruction-following diffusion, inversion, or additional spatial conditions~\cite{bartal2022text2live,hertz2022prompt,brooks2023instructpix2pix,mokady2023nulltext,zhang2023adding}. Built upon these image-editing priors, diffusion-based video editors either train dedicated video diffusion models for structure/content-guided editing~\cite{molad2023dreamix,esser2023structure} or adapt image diffusion models to videos through inversion, cross-frame attention, feature propagation, patch matching, canonical video representations, and image-to-video propagation~\cite{qi2023fatezero,liu2024videop2p,wang2023vid2vidzero,ceylan2023pix2video,yang2023rerender,geyer2024tokenflow,ouyang2024codef,ku2024anyv2v}. Related efforts also study controllable video editing, video inpainting, and diffusion-feature correspondence for editing consistency~\cite{ma2023magicstick,ma2025followcreation,wang2024cove}. Recent studies further investigate {RGBA} video generation, layer-wise alpha decomposition, and consistency learning for compositing or visually stable edits~\cite{chen2025transanimate,wang2025diffdecompose,song2025omniconsistency}. These methods have substantially improved prompt-based stylization, object replacement, and local attribute editing. Nevertheless, they are not specifically designed for production-level multi-character replacement: edits are often prompt-driven rather than identity-specific, temporal consistency may rely on short clips or propagation assumptions, and multiple independently controlled actors can easily suffer from identity leakage, foreground-background mismatch, or occlusion artifacts. In contrast, our task requires structure-guided recovery of multiple performers inside dynamic cinematic scenes, where the model must preserve background, camera motion, interactions, and lighting while rendering new actors consistently over long videos.

\section{Method}
\label{sec:method}

\begin{figure*}[!t]
\centering
\includegraphics[width=\linewidth]{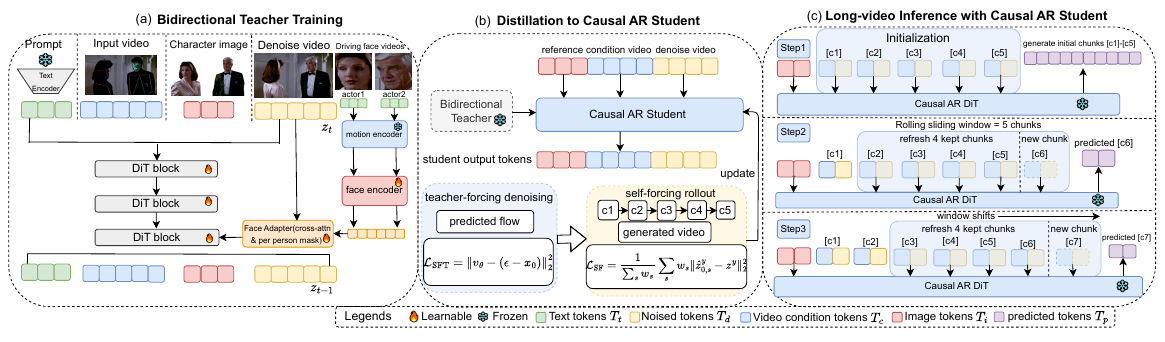}
\caption{Overview of PAI-Actor. (a) \textbf{Bidirectional Teacher}: Denoises target videos conditioned on text, control signals, and references. (b) \textbf{Causal Distillation}: A frozen teacher guides the AR student via two-stage teacher- and self-forcing rollout losses. (c) \textbf{Inference}: Autoregressive long-video generation using a 5-chunk sliding window.}
\label{fig:method}
\end{figure*}

We present \textbf{PAI-Actor}, a video-driven framework that preserves source motion and scene dynamics while rendering target actor appearances. As illustrated in Figure~\ref{fig:method}, our method follows a teacher-to-student design organized as follows: overall architecture formulation (Sec.~\ref{sec:overall_architecture}), a bidirectional in-context teacher for high-quality short clips (Sec.~\ref{sec:bidir_teacher}), a two-stage autoregressive distillation for scalable long-video generation (Sec.~\ref{sec:student_distillation}), KV-cache-based inference procedures (Sec.~\ref{sec:inference}), and a structure-guided training-data construction pipeline (Sec.~\ref{sec:data_engine}).

\subsection{Overall Architecture}
\label{sec:overall_architecture}

PAI-Actor treats cinematic character animation as a reference-conditioned video editing problem. 
The input video provides motion, expression, interaction, and scene dynamics, while the reference images specify the actor appearances. 
Using references from the source actor naturally supports video-driven reenactment; replacing them with references from another identity turns the same pipeline into character replacement.

The overall pipeline has two stages. 
First, we train a bidirectional diffusion transformer teacher with an in-context formulation. 
The teacher receives actor reference tokens, a rendered condition video, and noisy target-video tokens in one shared sequence, allowing full spatiotemporal attention across the short clip. 
This design is expensive in memory and computation, but is important for 1080P cinematic quality because the model can jointly reason about actor boundaries, background continuity, occlusion, and temporal context. 
In practice, full in-context training at 1080P is limited to short clips of about 97 frames, i.e., roughly four seconds, even on large-memory GPUs.

Second, we distill the bidirectional teacher into a causal autoregressive student. The primary motivation for this distillation is to overcome the temporal limitations of the bidirectional model and enable scalable long-form video generation. The student generates the video chunk by chunk, reuses previous chunks through a KV cache, and avoids computing full bidirectional attention over the entire sequence. Consequently, this autoregressive design provides the essential temporal scalability and computational memory efficiency required for practical cinematic production.

\subsection{Bidirectional In-Context Teacher}
\label{sec:bidir_teacher}

The teacher is trained as a structure-guided human recovery model. 
For each source video, we remove or degrade the original actor regions, and render body skeletons, face landmarks, and actor masks back onto the same video canvas. 
The resulting condition video preserves the original background and camera motion, while explicitly indicating where and how each actor should be reconstructed.

This condition design is deliberately video-like rather than a separate sparse control signal. 
By encoding the masked video, structural guidance, and scene content together, the model can better learn foreground-background fusion, occlusion handling, and boundary-consistent human recovery. 
This is especially important for cinematic scenes, where the actor is not isolated from the environment but interacts with lighting, shadows, camera motion, and other characters.

Actor references are encoded separately by a visual image encoder. 
For actor \(k\), the reference image \(r^k\) is mapped to appearance tokens \(\mathbf{a}^k\). 
The teacher receives the concatenated in-context sequence
\begin{equation}
\mathbf{Z}_{\mathrm{in}}
=
[\mathbf{a}^{1},\ldots,\mathbf{a}^{K} \mid \mathbf{z}^{c} \mid \mathbf{z}^{y}_{t}],
\end{equation}
where \(\mathbf{z}^{c}\) is the latent condition video and \(\mathbf{z}^{y}_{t}\) is the noisy target-video latent. 
We optimize the teacher with the rectified-flow objective:
\begin{equation}
\mathcal{L}_{\mathrm{teacher}}
=
\mathbb{E}
\left[
\left\|
F_{\phi}(\mathbf{z}^{y}_{t}, \mathbf{z}^{c}, \mathbf{A}, t)
-
\mathbf{v}^{*}
\right\|_{2}^{2}
\right],
\end{equation}
where \(\mathbf{A}=\{\mathbf{a}^{k}\}_{k=1}^{K}\) and \(\mathbf{v}^{*}\) is the target flow. 
Full bidirectional attention allows the teacher to use both past and future frames, which improves short-clip quality under occlusion, fast motion, and multi-character interaction.

Beyond the face landmarks rendered into the condition video, we inject each actor's facial expression through a dedicated \emph{face adapter}. For actor \(k\), the per-frame face crops \(f^{k}\) are passed through a frozen, pretrained motion encoder \(E_{\mathrm{mot}}\) to obtain an expression feature \(\mathbf{m}^{k}=E_{\mathrm{mot}}(f^{k})\), which a trainable face encoder \(E_{\mathrm{face}}\) maps to expression tokens \(\mathbf{e}^{k}=E_{\mathrm{face}}(\mathbf{m}^{k})\). Inside each adapter layer, the video hidden states \(\mathbf{h}\) attend to these tokens through per-actor masked cross-attention:
\begin{equation}
\mathbf{h} \;\leftarrow\; \mathbf{h} + \sum_{k=1}^{K} \mathbf{M}^{k}\odot
\mathrm{Attn}\!\left(\mathbf{h}\mathbf{W}_{Q},\; \mathbf{e}^{k}\mathbf{W}_{K},\; \mathbf{e}^{k}\mathbf{W}_{V}\right),
\end{equation}
where \(\mathbf{M}^{k}\) is the spatial mask of actor \(k\) that routes each expression signal only to that actor's own facial region, preventing cross-talk between characters. This dedicated pathway captures fine-grained facial motion that the rendered landmarks alone under-specify, and, as shown in our ablation (Table~\ref{tab:ablation_all_raw}), it markedly sharpens expression fidelity (Expr.-LMD) at negligible cost to the other metrics.

\subsection{Two-Stage Distillation for Long-Video Generation}
\label{sec:student_distillation}

The teacher produces high-quality short clips, but its full-attention in-context design cannot scale to long cinematic videos. 
We therefore convert it into a causal autoregressive student through two-stage distillation. 
Stage 1 adapts the architecture and timestep distribution for causal generation; Stage 2 trains the student on its own rollout distribution to reduce long-horizon drift.

\paragraph{Stage 1: Causal Adaptation.}
We initialize the student from the bidirectional teacher and replace full temporal attention with block-causal attention. 
The video is divided into temporal chunks. 
Tokens inside the current chunk can still attend bidirectionally to preserve local temporal quality, while chunk \(n\) can only attend to the reference tokens, condition tokens, and previous chunks. 
During inference, previous chunks are stored in a persistent KV cache, so the model can extend the video without recomputing the full history.

Causal generation also changes the noise structure. 
In autoregressive inference, past chunks are already clean and cached, while the current chunk is still noisy. 
To match this mixed state, Stage 1 samples independent timesteps per frame or per chunk, including clean states with \(t=0\). 
The student is trained with the same rectified-flow objective under the block-causal mask:
\begin{equation}
\mathcal{L}_{\mathrm{stage1}}
=
\mathbb{E}
\left[
\left\|
F_{\theta}(\mathbf{z}^{y}_{\mathbf{t}}, \mathbf{z}^{c}, \mathbf{A}, \mathbf{t})
-
\mathbf{v}^{*}_{\mathbf{t}}
\right\|_{2}^{2}
\right].
\end{equation}
This stage gives the student a causal architecture and exposes it to heterogeneous noise states, but the inputs are still derived from ground-truth latents.

\paragraph{Stage 2: On-Policy Self-Forcing.}
Stage 2 further aligns training with autoregressive inference.
Starting from the Stage-1 checkpoint, we roll out the student chunk by chunk over $S$ denoising steps with timesteps $t_{s}\in[\epsilon,1]$ ($\epsilon{=}10^{-3}$; the index $s$ is distinct from the actor index $k$ used above).
At step \(s\), the model predicts a clean latent \(\hat{\mathbf{z}}^{y}_{0,s}\), which is re-noised to form the next-step input:
\begin{equation}
\mathbf{z}^{y}_{t_{s+1}}
=
(1-t_{s+1})\hat{\mathbf{z}}^{y}_{0,s}
+
t_{s+1}\epsilon_{s+1}.
\end{equation}
Thus, the student is trained on inputs produced by itself, including KV caches built from its own previous predictions.
Since PAI-Actor is a conditional video-to-video task, each condition and reference set has a paired target video.
We can therefore directly supervise the rollout with the ground-truth latent:
\begin{equation}
\mathcal{L}_{\mathrm{stage2}}
=
\frac{1}{\sum_{s} w_{s}}
\sum_{s=1}^{S}
w_{s}
\left\|
\hat{\mathbf{z}}^{y}_{0,s}
-
\mathbf{z}^{y}
\right\|_{2}^{2},
\qquad
w_{s}=\frac{(1-t_{s})^{2}}{(t_{s}+\epsilon)^{2}}.
\end{equation}
The SNR weight (with the same small $\epsilon$ for numerical stability) emphasizes low-noise steps where the clean target is more reliable. 
We backpropagate through the re-noising trajectory, so early predictions are optimized not only to match the target, but also to produce intermediate states that later steps can refine. 
This reduces timestep drift and cache-induced chunk drift during long-video generation.

\subsection{Inference}
\label{sec:inference}

At inference time, we first convert the source video into a rendered condition video. 
We estimate actor masks, body skeletons, and face landmarks, dilate the masks by 5--20 pixels, remove the original actor regions, and render the structural signals into the editable regions. 
The mask dilation provides a boundary buffer, allowing the model to adapt to reference actors with different body contours, hairstyles, or clothing silhouettes.

The reference images are encoded into appearance tokens and concatenated with the condition-video tokens and noisy target tokens. 
The causal student then generates the output video autoregressively. 
For the first chunk, the model denoises from noise conditioned on the reference tokens and the first condition chunk. 
After each chunk is generated, we perform a clean \(t=0\) cache-writing pass to store its keys and values. 
Subsequent chunks attend to the global reference tokens, the current condition chunk, and 4 previously cached chunks, forming a sliding temporal window of 5 chunks in total. 

This inference process is the same for character animation and character replacement. 
The difference lies in the reference images: using references consistent with the source performer yields driven animation, while using references from a new identity yields character replacement. 
The bidirectional teacher can be used for highest-quality short clips, while the causal student is used for long-form generation.

\subsection{Training Data Construction}
\label{sec:data_engine}

We construct our training data from real-world movie footage. This ensures the model learns from natural character appearances, complex background dynamics, and authentic lighting conditions.

For a given original video sequence $\mathbf{V}$, we first extract the character masks $\mathbf{M}$ and the corresponding skeletal poses $\mathbf{P}$. To formulate the training condition, we mask out the character in the original video to create a black silhouette, effectively removing the original identity while perfectly preserving the background context. The colored skeletal pose is then spatially overlaid onto this masked sequence to construct the control video $\mathbf{V}_{ctrl}$. 
Formally, each training triplet is $\mathcal{T}=(I_{ref},V_{ctrl},V)$, where $I_{ref}$ provides the identity prior (typically sampled from the same video). The control video $V_{ctrl}=(1-M)\odot V\oplus P$ is constructed via mask-inverted element-wise multiplication ($\odot$) and spatial pose overlay ($\oplus$).

Crucially, during the extraction process, the character masks $\mathbf{M}$ are randomly dilated by 5--20 pixels before generating the control video. This dilation forces the model to learn how to seamlessly inpaint uncertain person-background boundaries and adapt the generated silhouette to novel reference identities with potentially different body shapes. By training on these real-world triplets $\mathcal{T}$, PAI-Actor learns robust, structure-guided character recovery under diverse and challenging cinematic conditions.
Because $\mathbf{V}_{ctrl}$ completely obscures the source identity, the model must learn to reconstruct the character conditioned entirely on $\mathbf{I}_{ref}$. This self-supervised formulation naturally enables zero-shot character replacement at inference simply by substituting $\mathbf{I}_{ref}$ with a novel identity.

\section{Experiments}
\label{sec:experiments}

\begin{figure*}[!t]
\animategraphics[width=\linewidth]{5}{fig/showcase/showcase_}{00}{09}
\caption{Generation results of PAI-Actor on two tasks. Readers can click and play the video clips in this figure using {\color{red}\textbf{Adobe Acrobat}}.}
\label{fig:results}
\end{figure*}

\subsection{Experimental Setup}
\label{sec:setup}

\paragraph{Implementation Details.}
Our model is trained on the proposed \textbf{PAI-Actor Dataset} (${\sim}$30K movie clips with diverse multi-character interactions, motions, and scenes), and evaluated on the \textbf{PAI-Benchmark}. During inference, outputs are generated at 1080P ($1920 \times 1056$) and 24 FPS, comprising 97 frames (${\sim}4$s) for short-clip and extended sequences for autoregressive testing using a temporal chunk size of 5 latent frames and a sliding window of 5 chunks. We finetune the base diffusion transformer in BF16 on 8 H200 GPUs (batch size 1) using LoRA~\citep{hu2022lora} (rank 80) and AdamW~\citep{loshchilov2019adamw} (lr=$5\times10^{-5}$). The bidirectional teacher is trained for ${\sim}8{,}000$ steps. The causal autoregressive student is then distilled via two stages: (i) teacher-forcing causal SFT (${\sim}1{,}600$ steps), and (ii) on-policy self-forcing with the frozen teacher (${\sim}950$ steps). Appendix~\ref{sec:suppl_cost} details wall-clock and memory costs.

\paragraph{Baseline Methods.}
We compare PAI-Actor against leading commercial platforms (\textbf{Kling O1}~\citep{kling}, \textbf{Domo}~\citep{domoai}, \textbf{Viggle}~\citep{viggle}) evaluated under their default APIs, and state-of-the-art open-source methods for controllable animation (\textbf{Wan-Animate}~\citep{wananimate}\footnote{Wan-Animate supports single-character task only; for multi-character task, we use it by replacing one character at a time and feeding the output as the next round's driving video. The corresponding rows are marked with $^{*}$ in Tables~\ref{tab:quantitative} and~\ref{tab:gemini}.}, \textbf{VACE}~\citep{vace}, \textbf{MoCha}~\citep{mocha}). Where applicable, we also include task-specific baselines (\textbf{MultiAnimate}~\citep{multianimate}, \textbf{SCAIL}~\citep{scail}, \textbf{SCAIL-2}) to comprehensively evaluate multi-character task and foreground-background fusion under comparable conditions. To ensure a fair comparison, all evaluated models are inferred under their respective optimal settings.

\begin{figure*}[!t]
\animategraphics[width=\linewidth]{5}{fig/compare/cmp_}{00}{09}
\caption{Comparison against Wan-Animate, MultiAnimate, SCAIL, SCAIL-2, DOMO, Viggle, and Kling O1. Readers can click and play the video clips in this figure using {\color{red}\textbf{Adobe Acrobat}}.}
\label{fig:comp2}
\end{figure*}

\paragraph{Benchmark.}
We construct the \textbf{PAI-Benchmark} comprising three subsets:
(1) \textbf{Character-Animation Subset} evaluates self-reconstruction on 100 movie clips (disjoint from training). The reference image matches the source actor, providing natural ground-truth to measure motion and appearance fidelity.
(2) \textbf{Character-Replacement Subset} evaluates zero-shot generalization on 100 unseen movie clips. Here, source actors are replaced with novel reference identities, testing the model's robustness under complex multi-character interactions where no ground-truth exists. For each target identity, we first generate a full-body standing portrait and use Gemini~2.5 Flash Image~\citep{nanobanana} to re-pose it into a configuration approximating a representative pose of the corresponding character in the source video. To ensure a strictly fair comparison, these \textbf{co-posed} images are provided as the standard reference input for all evaluated models, including all baselines. (3) \textbf{Long-Video Subset} evaluates temporal scalability using 40 movie clips (20 single-character and 20 multi-character), consisting of long-form sequences of $\sim$417 frames ($\sim$17\,s at 24 FPS) per clip.

\textbf{Evaluation Metrics.} We evaluate along four dimensions:
(1) \textbf{Reconstruction Quality:} On the Character-Animation Subset, we report full-frame MSE, SSIM, and LPIPS~\citep{lpips} against the ground truth. For baselines that intentionally alter backgrounds (MultiAnimate, SCAIL), we compute metrics exclusively on masked character regions to ensure fair comparison.
(2) \textbf{Motion Fidelity:} We extract whole-body keypoints (DWPose~\citep{dwpose} fused with NLF-pose~\citep{nlfpose}) to compute \textit{Pose-Dist Mean} and \textit{Pose-Match}. We also use CoTracker3~\citep{cotracker3} to track dense point grids, reporting the cosine similarity of displacement trajectories (\textit{Motion Cons.}).
(3) \textbf{Facial Expression Fidelity:} We report the \emph{Expression Landmark Distance} (\textit{Expr.-LMD}). From the $68$ DWPose~\citep{dwpose} facial keypoints, we estimate a rigid transform from the expression-invariant eye corners and nose ridge and apply it to the whole face (removing head pose and scale), so every landmark is brought into a common frame; we then average the distance between corresponding mouth, eye, brow, and nose landmarks, normalized by the inter-ocular distance.
(4) \textbf{LLM-based Perceptual Evaluation:} GPT-5.1~\citep{gpt5} and Gemini-3.1-Pro~\citep{gemini} score \textit{BG Cons.}, \textit{FG-BG Fusion}, \textit{Illum. Harmony}, and \textit{Char. Cons.} on a 1--10 scale. Final scores average both judges' 3-query medians, demonstrating strong agreement (mean Spearman $\rho=0.75$).

\subsection{Results}
\label{sec:results}

The generation results are shown in Figure~\ref{fig:results}. PAI-Actor demonstrates strong temporal stability and identity consistency across frames, even in complex multi-character scenarios. Unlike existing approaches that suffer from identity drift or inconsistent rendering, our method maintains precise character appearance while faithfully reproducing motion dynamics. Our model also can generate diverse character types across various scenes, including anime, aliens, and realistic humans. Additional results are provided in the \textbf{appendix}.

Furthermore, our framework scales effectively to multi-character settings. Multiple characters can be independently controlled within the same scene without identity mixing or structural artifacts. This is particularly evident in interaction-heavy scenarios, where baseline methods often fail.

Finally, the autoregressive distillation design enables long-form video generation with stable temporal coherence. Compared to diffusion-based baselines limited to short clips, PAI-Actor produces extended sequences without noticeable degradation on background and motion consistency.

\begin{table*}[!t]
\centering
\caption{Quantitative comparison on the Character-Animation Subset under both single-character and multi-character settings. BG Cons., FG-BG Fusion, Illum. Harmony, and Char. Cons. are evaluated via LLM-based perceptual scoring.}
\label{tab:quantitative}
\renewcommand{\arraystretch}{1.0}
\setlength{\tabcolsep}{4pt}
\resizebox{\textwidth}{!}{\footnotesize
\begin{tabular}{l*{22}{c}}
\toprule
\multirow{2}{*}{\textbf{Method}}
& \multicolumn{2}{c}{\shortstack{\textbf{MSE} $\downarrow$ \\ \scriptsize$(\times 10^{-3})$}}
& \multicolumn{2}{c}{\textbf{SSIM} $\uparrow$}
& \multicolumn{2}{c}{\textbf{LPIPS} $\downarrow$}
& \multicolumn{2}{c}{\shortstack{\textbf{Pose-Dist} $\downarrow$ \\ \scriptsize$(\times 10^{-2})$}}
& \multicolumn{2}{c}{\shortstack{\textbf{Pose-Match} $\uparrow$}}
& \multicolumn{2}{c}{\shortstack{\textbf{Motion} \\ \textbf{Cons.} $\uparrow$}}
& \multicolumn{2}{c}{\shortstack{\textbf{BG} \\ \textbf{Cons.} $\uparrow$}}
& \multicolumn{2}{c}{\shortstack{\textbf{FG-BG} \\ \textbf{Fusion} $\uparrow$}}
& \multicolumn{2}{c}{\shortstack{\textbf{Illum.} \\ \textbf{Harmony} $\uparrow$}}
& \multicolumn{2}{c}{\shortstack{\textbf{Char.} \\ \textbf{Cons.} $\uparrow$}}
& \multicolumn{2}{c}{\shortstack{\textbf{Expr.-LMD} $\downarrow$ \\ \scriptsize$(\times 10^{-2})$}} \\
\cmidrule(lr){2-3} \cmidrule(lr){4-5} \cmidrule(lr){6-7} \cmidrule(lr){8-9} \cmidrule(lr){10-11} \cmidrule(lr){12-13} \cmidrule(lr){14-15} \cmidrule(lr){16-17} \cmidrule(lr){18-19} \cmidrule(lr){20-21} \cmidrule(lr){22-23}
& {\scriptsize\textbf{Single}} & {\scriptsize\textbf{Multi}}
& {\scriptsize\textbf{Single}} & {\scriptsize\textbf{Multi}}
& {\scriptsize\textbf{Single}} & {\scriptsize\textbf{Multi}}
& {\scriptsize\textbf{Single}} & {\scriptsize\textbf{Multi}}
& {\scriptsize\textbf{Single}} & {\scriptsize\textbf{Multi}}
& {\scriptsize\textbf{Single}} & {\scriptsize\textbf{Multi}}
& {\scriptsize\textbf{Single}} & {\scriptsize\textbf{Multi}}
& {\scriptsize\textbf{Single}} & {\scriptsize\textbf{Multi}}
& {\scriptsize\textbf{Single}} & {\scriptsize\textbf{Multi}}
& {\scriptsize\textbf{Single}} & {\scriptsize\textbf{Multi}}
& {\scriptsize\textbf{Single}} & {\scriptsize\textbf{Multi}} \\
\midrule
Domo     & 17.0          & 19.1          & 0.647 & 0.623 & 0.335 & 0.360 & 4.65          & 4.09          & 0.674 & 0.857 & 0.732 & 0.791 & 8.24 & 8.04 & 7.02 & 6.78 & 7.19 & 7.52 & \textbf{5.67} & 4.59 & 7.39 & 19.41 \\
Viggle   & 12.8          & 16.6          & 0.718 & 0.700 & 0.263 & 0.312 & 4.66          & 5.18          & 0.637 & 0.710 & 0.627 & 0.714 & 8.10 & 8.13 & 5.40 & 4.89 & 6.17 & 5.85 & 3.90 & 3.39 & 5.71 & 15.07 \\
Kling O1 & 7.88           & 8.85           & 0.712 & 0.733 & 0.211 & 0.227 & 3.12          & 2.79          & 0.807 & 0.910 & 0.716 & 0.759 & 7.86 & 7.65 & 7.78 & 7.71 & 7.32 & 7.81 & 5.63 & \textbf{5.53} & 3.30 & 6.58 \\
Wan-Animate$^{*}$     & 5.30           & 9.94           & 0.832 & 0.766 & 0.170 & 0.251 & 2.65          & 3.21          & 0.833 & 0.898 & 0.796 & 0.814 & 8.05 & 8.12 & 7.85 & 7.57 & 7.50 & 7.43 & 4.90 & 4.32 & 2.36 & 12.09 \\
SCAIL                 & 52.9          & 47.6          & 0.852 & 0.779 & 0.161 & 0.217 & 4.74          & 4.00          & 0.755 & 0.835 & 0.749 & 0.736 & 2.59 & 2.64 & 5.46 & 5.64 & 6.49 & 6.63 & 5.07 & 4.47 & 4.68 & 22.09 \\
SCAIL-2               & 79.0          & 98.2          & 0.809 & 0.739 & 0.207 & 0.294 & 5.97          & 5.80          & 0.557 & 0.648 & 0.697 & 0.627 & 2.72 & 2.44 & 7.06 & 6.25 & 7.09 & 6.92 & 4.87 & 2.86 & 3.38 & 13.49 \\
MultiAnimate         & 26.2          & 32.6          & \textbf{0.905} & \textbf{0.868} & 0.126 & 0.170 & 3.78          & 3.06          & 0.766 & 0.891 & 0.524 & 0.518 & 2.57 & 2.32 & 5.17 & 4.63 & 6.52 & 6.02 & 5.27 & 5.19 & 3.77 & 22.39 \\
VACE                  & 3.22           & 4.77           & 0.833 & 0.836 & 0.169 & 0.188 & 2.63          & 2.63          & 0.862 & 0.907 & 0.751 & 0.774 & 8.33 & 8.37 & 7.83 & 7.63 & \textbf{7.83} & 7.57 & 5.17 & 5.07 & 3.66 & 16.16 \\
\midrule
\rowcolor{highlightcolor}
\textbf{Ours-Bidir}
& \textbf{1.89}  & \textbf{2.31}
& 0.856 & 0.861
& \textbf{0.114} & \textbf{0.123}
& \textbf{2.30}  & \textbf{2.08}
& \textbf{0.871} & \textbf{0.946}
& \textbf{0.813} & \textbf{0.850}
& \textbf{8.41} & \textbf{8.48} 
& \textbf{7.95} & \textbf{8.16} 
& 7.71 & \textbf{7.89} 
& 5.50 & 5.32 
& \textbf{2.15}  & \textbf{4.97} \\
\bottomrule
\end{tabular}}
\end{table*}

\begin{table*}[!t]
\centering
\caption{Quantitative results of \texttt{Ours-Causal} on the Long-Video Subset.}
\label{tab:longvideo}
\renewcommand{\arraystretch}{1.0}
\setlength{\tabcolsep}{4pt}
\resizebox{\textwidth}{!}{\footnotesize
\begin{tabular}{l*{20}{c}}
\toprule
\multirow{2}{*}{\textbf{Method}}
& \multicolumn{2}{c}{\shortstack{\textbf{MSE} $\downarrow$ \\ \scriptsize$(\times 10^{-3})$}}
& \multicolumn{2}{c}{\textbf{SSIM} $\uparrow$}
& \multicolumn{2}{c}{\textbf{LPIPS} $\downarrow$}
& \multicolumn{2}{c}{\shortstack{\textbf{Pose-Dist} $\downarrow$ \\ \scriptsize$(\times 10^{-2})$}}
& \multicolumn{2}{c}{\shortstack{\textbf{Pose-Match} $\uparrow$}}
& \multicolumn{2}{c}{\shortstack{\textbf{Motion} \\ \textbf{Cons.} $\uparrow$}}
& \multicolumn{2}{c}{\shortstack{\textbf{BG} \\ \textbf{Cons.} $\uparrow$}}
& \multicolumn{2}{c}{\shortstack{\textbf{FG-BG} \\ \textbf{Fusion} $\uparrow$}}
& \multicolumn{2}{c}{\shortstack{\textbf{Illum.} \\ \textbf{Harmony} $\uparrow$}}
& \multicolumn{2}{c}{\shortstack{\textbf{Char.} \\ \textbf{Cons.} $\uparrow$}} \\
\cmidrule(lr){2-3} \cmidrule(lr){4-5} \cmidrule(lr){6-7} \cmidrule(lr){8-9} \cmidrule(lr){10-11} \cmidrule(lr){12-13} \cmidrule(lr){14-15} \cmidrule(lr){16-17} \cmidrule(lr){18-19} \cmidrule(lr){20-21}
& {\scriptsize\textbf{Single}} & {\scriptsize\textbf{Multi}}
& {\scriptsize\textbf{Single}} & {\scriptsize\textbf{Multi}}
& {\scriptsize\textbf{Single}} & {\scriptsize\textbf{Multi}}
& {\scriptsize\textbf{Single}} & {\scriptsize\textbf{Multi}}
& {\scriptsize\textbf{Single}} & {\scriptsize\textbf{Multi}}
& {\scriptsize\textbf{Single}} & {\scriptsize\textbf{Multi}}
& {\scriptsize\textbf{Single}} & {\scriptsize\textbf{Multi}}
& {\scriptsize\textbf{Single}} & {\scriptsize\textbf{Multi}}
& {\scriptsize\textbf{Single}} & {\scriptsize\textbf{Multi}}
& {\scriptsize\textbf{Single}} & {\scriptsize\textbf{Multi}} \\
\midrule
\rowcolor{highlightcolor}
\textbf{Ours-Causal}
& 4.77  & 4.48
& 0.880 & 0.860
& 0.118 & 0.129
& 4.90  & 1.56
& 0.857 & 0.988
& 0.856 & 0.903
& 9.88  & 9.43
& 6.25  & 6.22
& 7.25  & 7.97
& 5.38  & 4.33 \\
\bottomrule
\end{tabular}}
\end{table*}

\begin{table}[!t]
\centering
\caption{LLM-based evaluation (averaged GPT-5.1 \& Gemini-3.1-Pro) on the Character-Replacement Subset.}
\label{tab:gemini}
\renewcommand{\arraystretch}{0.7}
\setlength{\tabcolsep}{3pt}
\footnotesize
\resizebox{\columnwidth}{!}{%
\begin{tabular}{l*{8}{c}}
\toprule
\multirow{2}{*}{\textbf{Method}}
& \multicolumn{2}{c}{\shortstack{\textbf{BG} \\ \textbf{Cons.} $\uparrow$}}
& \multicolumn{2}{c}{\shortstack{\textbf{FG-BG} \\ \textbf{Fusion} $\uparrow$}}
& \multicolumn{2}{c}{\shortstack{\textbf{Illum.} \\ \textbf{Harmony} $\uparrow$}}
& \multicolumn{2}{c}{\shortstack{\textbf{Char.} \\ \textbf{Cons.} $\uparrow$}} \\
\cmidrule(lr){2-3} \cmidrule(lr){4-5} \cmidrule(lr){6-7} \cmidrule(lr){8-9}
& {\scriptsize\textbf{Single}} & {\scriptsize\textbf{Multi}}
& {\scriptsize\textbf{Single}} & {\scriptsize\textbf{Multi}}
& {\scriptsize\textbf{Single}} & {\scriptsize\textbf{Multi}}
& {\scriptsize\textbf{Single}} & {\scriptsize\textbf{Multi}} \\
\midrule
Domo              & 8.04 & 8.02 & 6.94 & 6.78 & 6.98 & 7.46 & 6.33 & 5.78 \\
Viggle            & 7.56 & 7.74 & 5.15 & 4.04 & 6.07 & 5.76 & 5.22 & 4.24 \\
Kling O1          & 7.76 & 7.69 & 7.41 & 7.63 & 7.26 & 7.44 & 6.92 & 6.45 \\
Wan-Animate$^{*}$ & 7.89 & 7.85 & 7.48 & 7.28 & \textbf{7.49} & 7.50 & 6.36 & 5.04 \\
MoCha             & 7.92 & - & 7.44 & - & 7.38 & - & 5.83 & - \\
SCAIL-2           & 6.80 & 6.53 & 7.15 & 6.45 & 6.89 & 7.19 & 6.60 & 4.05 \\
\midrule
\rowcolor{highlightcolor}
\textbf{Ours-Bidir}     & \textbf{8.36} & \textbf{8.36} & \textbf{7.54} & \textbf{7.82} & 7.29 & \textbf{7.63} & \textbf{6.95} & \textbf{6.46} \\
\bottomrule
\end{tabular}}
\end{table}

\subsection{Comparison and Evaluation}
\label{sec:comparison}

\paragraph{Quantitative Evaluation.}

Tables~\ref{tab:quantitative} and \ref{tab:gemini} report quantitative comparisons. PAI-Actor achieves state-of-the-art performance across the majority of metrics. While MultiAnimate obtains higher SSIM due to an easier foreground-only evaluation (necessitated by its background-altering task objective), our method is strictly evaluated on the full frame. Despite this harder setting, PAI-Actor maintains highly competitive SSIM and consistently outperforms all baselines in motion fidelity, background consistency, and multi-character identity preservation. Furthermore, on the Long-Video Subset (Table~\ref{tab:longvideo}), our \texttt{Ours-Causal} student maintains competitive performance even on $\sim$417-frame sequences ($\sim$17\,s at 24\,FPS), demonstrating that the proposed bidirectional-to-autoregressive distillation transfers cinematic-quality short-clip generation to scalable long-form video synthesis.

In particular, strong identity consistency scores demonstrate that our method effectively preserves character appearance across frames. Motion fidelity results confirm accurate transfer of motion and expressions. Moreover, superior performance in multi-character consistency validates the effectiveness of our unified in-context formulation.

\paragraph{Qualitative Evaluation.}
Figure~\ref{fig:comp2} presents qualitative comparisons for the Character-Animation and Character-Replacement Subset. Existing methods commonly suffer from motion misalignment, background degradation, and identity drift. In contrast, PAI-Actor accurately adheres to the driving pose while maintaining strict identity consistency and preserving the original background context.

Notably, our method excels in highly challenging scenarios where baselines typically fail. It robustly handles complex occlusions and seamlessly retains unmasked handheld objects, which are often erroneously erased by other approaches. Furthermore, PAI-Actor successfully captures complex physical effects from the source video, naturally integrating the generated character with dynamic lighting conditions and volumetric light, resulting in highly realistic and immersive outputs.

\subsection{Ablation Study}

\label{sec:ablation}

We conduct ablation experiments on the Character-Animation subset (Table~\ref{tab:ablation_all_raw}), where the full \texttt{Ours-Bidir} achieves the best or near-best performance across metrics. Qualitatively (Figure~\ref{fig:ablation}), dropping the co-posed reference (\texttt{w/o Co-Posed Ref.}) deprives the model of aligned spatial guidance; as shown in (a), it struggles to infer the correct body orientation, resulting in pose misalignment. Second, removing mask dilation (\texttt{w/o Dilation}) restricts generation to the source silhouette, causing boundary artifacts such as unnaturally cropped hair (b). Finally, the comparison in (c) reveals that the purely causal model prior to distillation (\texttt{Ours-Causal w/o self-forcing}) can occasionally suffer from noticeable facial degradation, whereas the full \texttt{Ours-Bidir} maintains better visual consistency. Finally, adding the face-encoder adapter yields the best Expr.-LMD ($2.15$/$4.97$ vs.\ $2.81$/$5.08$ without it), confirming that it sharpens facial-expression fidelity.

\begin{figure*}[!t]
\animategraphics[width=\linewidth]{5}{fig/ablation/abl_}{00}{09}
\caption{Ablation study on the Character-Animation Subset. The rightmost column is our full model. Readers can click and play the video clips in this figure using {\color{red}\textbf{Adobe Acrobat}}.}
\label{fig:ablation}
\end{figure*}

\begin{table*}[!t]
\centering
\caption{Ablation study on the Character-Animation subset. Each component was evaluated under single-character and multi-character settings.}
\label{tab:ablation_all_raw}
\renewcommand{\arraystretch}{1.0}
\setlength{\tabcolsep}{8pt}
\resizebox{\textwidth}{!}{\footnotesize
\begin{tabular}{l*{14}{c}}
\toprule
\multirow{2}{*}{\textbf{Method}}
& \multicolumn{2}{c}{\shortstack{\textbf{MSE} $\downarrow$ \\ \scriptsize$(\times 10^{-3})$}}
& \multicolumn{2}{c}{\textbf{SSIM} $\uparrow$}
& \multicolumn{2}{c}{\textbf{LPIPS} $\downarrow$}
& \multicolumn{2}{c}{\shortstack{\textbf{Pose-Dist} $\downarrow$ \\ \scriptsize$(\times 10^{-2})$}}
& \multicolumn{2}{c}{\textbf{Pose-Match} $\uparrow$}
& \multicolumn{2}{c}{\shortstack{\textbf{Motion} \\ \textbf{Cons.} $\uparrow$}}
& \multicolumn{2}{c}{\shortstack{\textbf{Expr.-LMD} $\downarrow$ \\ \scriptsize$(\times 10^{-2})$}} \\
\cmidrule(lr){2-3} \cmidrule(lr){4-5} \cmidrule(lr){6-7} \cmidrule(lr){8-9} \cmidrule(lr){10-11} \cmidrule(lr){12-13} \cmidrule(lr){14-15}
& {\scriptsize\textbf{Single}} & {\scriptsize\textbf{Multi}}
& {\scriptsize\textbf{Single}} & {\scriptsize\textbf{Multi}}
& {\scriptsize\textbf{Single}} & {\scriptsize\textbf{Multi}}
& {\scriptsize\textbf{Single}} & {\scriptsize\textbf{Multi}}
& {\scriptsize\textbf{Single}} & {\scriptsize\textbf{Multi}}
& {\scriptsize\textbf{Single}} & {\scriptsize\textbf{Multi}}
& {\scriptsize\textbf{Single}} & {\scriptsize\textbf{Multi}} \\
\midrule
Ours-Bidir w/o Dilation
& 2.36 & 3.14 & 0.830 & 0.834 & 0.117 & 0.130 & 2.48 & 2.29 & 0.862 & 0.914 & 0.809 & 0.847 & 2.95 & 5.69 \\
Ours-Bidir w/o Co-Posed Ref.
& 6.24 & 14.9 & 0.812 & 0.765 & 0.164 & 0.218 & 2.99 & 3.39 & 0.804 & 0.848 & 0.816 & 0.824 & 4.34 & 6.42 \\
Ours-Causal w/o self-forcing
& 2.76 & 3.84 & \textbf{0.878} & 0.860 & 0.121 & 0.139 & 2.73 & 2.35 & 0.819 & 0.917 & \textbf{0.826} & 0.842 & 4.16 & 5.76 \\
Ours-Bidir w/o adapter
& 1.91 & 2.32 & 0.855 & 0.861 & 0.115 & 0.124 & \textbf{2.28} & \textbf{2.07} & 0.843 & 0.923 & 0.814 & 0.847 & 2.81 & 5.08 \\
\midrule
\rowcolor{highlightcolor}
Ours-Bidir
& \textbf{1.89} & \textbf{2.31} & 0.856 & \textbf{0.861} & \textbf{0.114} & \textbf{0.123} & 2.30 & 2.08 & \textbf{0.871} & \textbf{0.946} & 0.813 & \textbf{0.850} & \textbf{2.15} & \textbf{4.97} \\
\bottomrule
\end{tabular}}
\end{table*}


    
    

\section{Conclusion}

In this work, we present \textbf{PAI-Actor}, a cinematic video-driven framework for multi-character animation and character replacement in dynamic scenes. Unlike prior systems limited to single-character or short-clip generation, PAI-Actor targets demanding film-level settings: preserving source motion, actor interactions, camera trajectories, and background dynamics while rendering target identities at 1080P resolution. 
We achieve this via a structure-guided data engine from real movie footage. We train a bidirectional diffusion teacher for high-quality short clips, which is then distilled into a causal autoregressive student to enable scalable, KV-cache-efficient long-form generation.
Built upon the open-source Wan2.2-TI2V-5B backbone, extensive evaluations show that PAI-Actor surpasses both recent open-source approaches and commercial systems in motion fidelity, identity consistency, background preservation, and multi-character controllability. These results demonstrate that high-quality cinematic character replacement is achievable through real-world structure-guided recovery, reference-conditioned in-context modeling, and long-video autoregressive distillation.
{
    \small
    \bibliographystyle{ieeenat_fullname}
    \bibliography{main}
}

\clearpage

\appendix

\newpage
\begin{center}
  {\Large\bfseries Appendix for\\[2pt]
   PAI-Actor: Cinematic Multi-Character Replacement\\
   in Dynamic Scenes\par}
\end{center}

The appendix is organized as follows:
\begin{enumerate}
    \item Sec.~\ref{sec:suppl_user_study}: details of the user study, including the interface, evaluation criteria, and randomization.
    \item Sec.~\ref{sec:suppl_llm}: prompt templates and metric definitions for our automated evaluation on the Character-Animation and Character-Replacement Subset.
    \item Sec.~\ref{sec:suppl_cotracker}: protocol and the scale-invariant trajectory cosine similarity used to measure motion fidelity with CoTracker3.
    \item Sec.~\ref{sec:suppl_correlation}: statistical correlation between the automated scores and human preference.
    \item Sec.~\ref{sec:suppl_cost}: computational cost and resources for training and inference.
    \item Sec.~\ref{sec:more_results}: additional results and visualizations.
    \item Sec.~\ref{sec:limitations}: limitations, future work, and analysis of typical failure cases.
\end{enumerate}

\section{User Study Details}
\label{sec:suppl_user_study}

We conduct a perceptual user study to compare PAI-Actor against representative open-source and commercial baselines. The study is organized into four tasks that jointly cover both directions of our setting and both single- and multi-character regimes:
\begin{itemize}
    \item \textbf{single-character} $\times$ \textbf{character-animation},
    \item \textbf{single-character} $\times$ \textbf{character-replacement},
    \item \textbf{multi-character} $\times$ \textbf{character-animation},
    \item \textbf{multi-character} $\times$ \textbf{character-replacement}.
\end{itemize}
Each task contains $10$ video groups, giving $40$ groups per participant in total.

\textbf{Interface.}
For each group, the actor reference image and the source performance video are displayed at the top of the page. Below them, the outputs from all compared methods are shown side-by-side as anonymized videos labelled A, B, C, \ldots. The order of the methods is randomized so that participants do not know which video belongs to which method. Synchronized playback controls (\textit{Play All}, \textit{Pause All}, \textit{Replay All}) are provided so that participants can compare temporal consistency across methods.

\textbf{Evaluation Criteria.}
For each group, participants answer four single-choice questions, each asking them to pick the single best video for one criterion:
\begin{enumerate}
    \item \textbf{Overall Preference}: which video is the most realistic and visually pleasing overall, given the reference video and the actor reference image.
    \item \textbf{Motion Fidelity}: which video best reproduces the body pose, gestures, and dynamics of the reference video.
    \item \textbf{Character Identity}: which video best preserves the actor's face, clothing, and overall appearance from the reference image.
    \item \textbf{Background Consistency}: which video most faithfully preserves the background of the reference video.
\end{enumerate}

\begin{figure*}[!t]
\centering
\includegraphics[width=\linewidth]{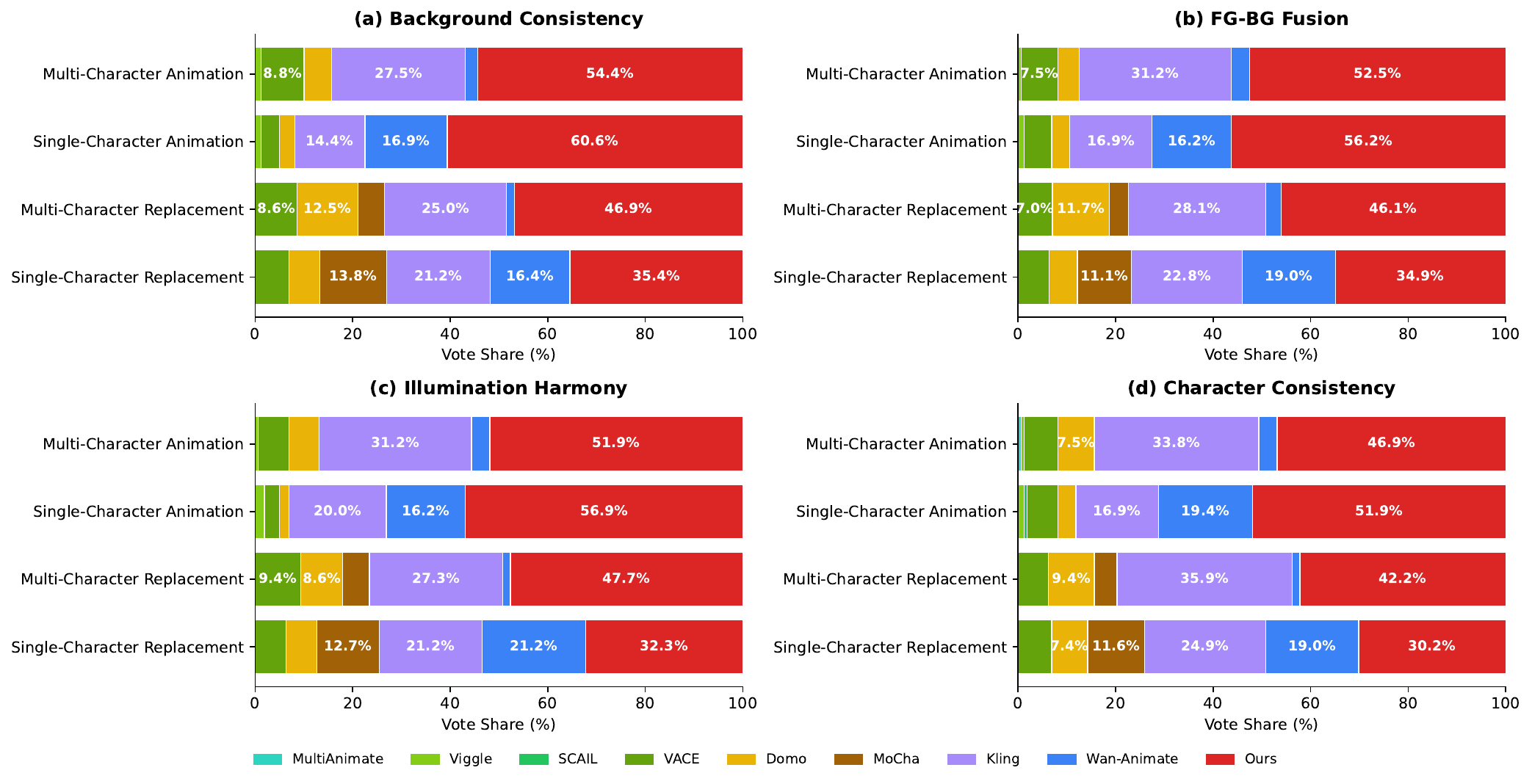}
\caption{User study results across the four perceptual criteria (\textbf{Background Consistency}, \textbf{FG-BG Fusion}, \textbf{Illumination Harmony}, \textbf{Character Consistency}). Each subplot shows the vote share (\%) per method on the four task settings (single/multi $\times$ animation/replacement). Values are aggregated over the $16$ qualified participants. Our method achieves the largest vote share on every (task, criterion) cell.}
\label{fig:user-study}
\end{figure*}

\textbf{Participants and Statistical Analysis.}
A total of $16$ participants with prior experience in video editing or visual content creation completed all $40$ groups, yielding $160$ evaluation choices per questionnaire across the four perceptual criteria. Aggregated per-method win rates across the four task settings $\times$ four perceptual criteria are visualized in Figure~\ref{fig:user-study}. Our method achieves the highest vote share in every (task, criterion) cell, confirming consistent human preference across all settings and dimensions. Statistical agreement between user-study votes and our LLM-based perceptual scores is reported in Section~\ref{sec:suppl_correlation} (Table~\ref{tab:correlation}), with average Spearman~$\rho=0.554$ on single-character and $\rho=0.724$ on multi-character settings (all $p<0.01$), confirming that our automated evaluation protocol agrees with human consensus.

\begin{figure}[!t]
    \centering
    \includegraphics[width=\linewidth]{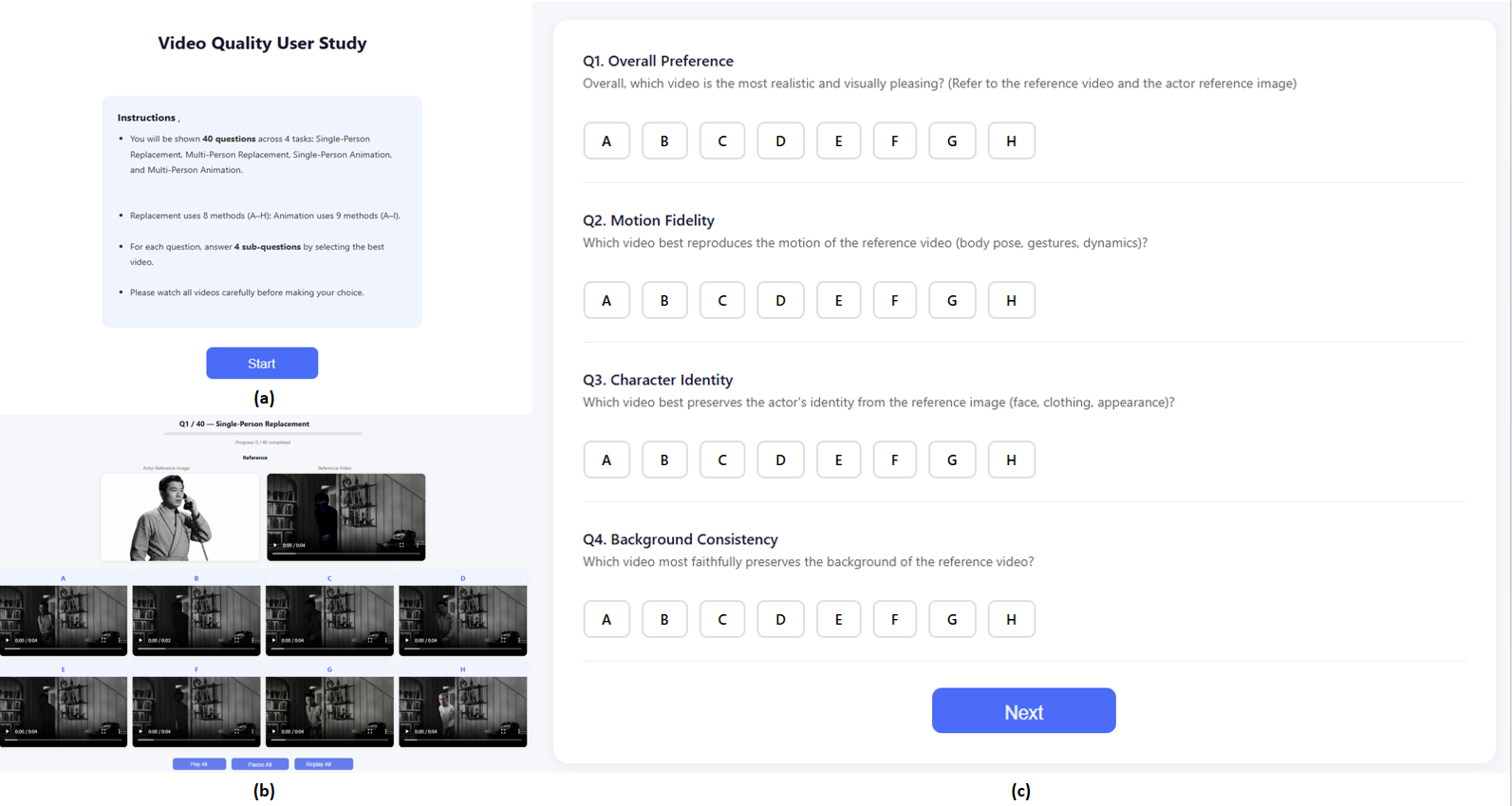}
    \caption{\textbf{User study interface and workflow.} (a) Study instructions, (b) side-by-side presentation of a test case with input references and anonymized results, and (c) the questionnaire with four evaluation criteria.}
    \label{fig:user_study_ui}
\end{figure}
\section{LLM-based Evaluation Details}
\label{sec:suppl_llm}

We employ two strong multimodal large language models, GPT-5.1~\citep{gpt5} and Gemini-3.1-Pro~\citep{gemini}, as automated perceptual judges for both subsets. To reduce single-judge bias and scoring variance, each metric is queried three times per model to obtain a median score, and the final reported score is the average of the medians from both GPT-5.1 and Gemini-3.1-Pro.

\textbf{Inputs.}
For every video, the judges receive (i)~the source performance video, (ii)~the generated video, and (iii)~the actor reference image(s). They are asked to return JSON of the form \texttt{\{``score'': <1--10>, ``reason'': ``<sentence>''\}} for each of four metrics: \textbf{Background Consistency} (\texttt{BG\_Cons}), \textbf{Foreground--Background Fusion} (\texttt{FG-BG\_Fusion}), \textbf{Illumination Harmony} (\texttt{Illum\_Harmony}), and \textbf{Character Consistency} (\texttt{Char\_Cons}). The full prompt templates used for both judges are shown in Figure~\ref{fig:prompt_llm_part1} and Figure~\ref{fig:prompt_llm_part2}. Aggregated results are reported in Table~\ref{tab:quantitative} and Table~\ref{tab:gemini} of the main paper.

\tcbset{
  promptbox/.style={
    colback=blue!3, colframe=blue!50!black,
    fonttitle=\bfseries\small, coltitle=white, colbacktitle=blue!55!black,
    boxrule=0.5pt, arc=2pt, left=4pt, right=4pt, top=2pt, bottom=2pt,
    before skip=4pt, after skip=4pt,
  },
  metricbox/.style={
    colback=gray!5, colframe=gray!50,
    boxrule=0.4pt, arc=1.5pt, left=3pt, right=3pt, top=2pt, bottom=2pt,
    before skip=3pt, after skip=3pt,
  }
}

\begin{figure*}[!t]
\begin{tcolorbox}[promptbox, title=LLM Evaluation Prompts (Part 1/2): BG\_Cons \& FG-BG\_Fusion]
\footnotesize

\textbf{Common Setup.}
\textit{Inputs:} For each metric, the judge receives a sequence of frames sampled uniformly from the generated video and (when applicable) the source / reference. \textit{Output:} JSON \texttt{\{``score'': <1--10>, ``reason'': ``<sentence>''\}}.

\begin{tcolorbox}[metricbox]
\textbf{Background Consistency} (\texttt{BG\_Cons}) \hfill \textit{Role: Expert evaluator for background preservation.}\\[2pt]
\textit{Inputs:} (1)~Control sequence: frames from the source video where the foreground subject(s) are replaced with a black silhouette overlaid with colored skeleton lines; (2)~Generated sequence: frames from the method output (foreground is the rendered character).\\[1pt]
\textit{Task:} Judge how consistent the generated background is compared to the source video's background. The judge must \emph{ignore the foreground in both sequences} (the silhouette+skeleton on the control side and the rendered character on the generated side) and only evaluate non-foreground background regions.\\[1pt]
\textit{Criteria:}
\begin{itemize}[nosep,leftmargin=12pt,label=\textbullet]
\item Scene identity: same room / street / environment as the control
\item Structure \& layout: spatial relationships, furniture, walls, no missing or added elements
\item Motion \& camera: same camera pan/tilt/parallax as the control
\item Temporal consistency: stable across frames, no flicker, warping, or texture crawling
\end{itemize}
\textit{Acceptable (do not penalize):} minor compression artifacts at mask boundaries; global color/lighting/tone shifts.\\[1pt]
\textit{Scoring:}
10: virtually identical background $|$
8--9: very faithful, minor structural diff $|$
6--7: clearly same scene, some changes $|$
4--5: visible structural differences $|$
2--3: major changes (missing/added objects) $|$
1: unrecognizable.
\end{tcolorbox}

\begin{tcolorbox}[metricbox]
\textbf{Foreground--Background Fusion} (\texttt{FG-BG\_Fusion}) \hfill \textit{Role: Expert evaluator for compositing realism.}\\[2pt]
\textit{Input:} A sequence of frames sampled uniformly from one generated video.\\[1pt]
\textit{Task:} Evaluate foreground--background fusion realism (anterior--posterior fusion).\\[1pt]
\textit{Criteria:}
\begin{itemize}[nosep,leftmargin=12pt,label=\textbullet]
\item Scale \& perspective: subject fits the scene geometry and camera viewpoint
\item Contact \& depth cues: feet contact ground; occlusion boundaries look correct
\item Shadows / reflections: consistent with background and subject placement
\item Color \& tone match: foreground grading matches background (contrast, saturation, grain, sharpness)
\item Temporal integration: no floating, jitter, or background leaking onto the subject
\end{itemize}
\textit{Scoring:}
10: seamless, looks like one real shot $|$
8--9: very good, minor grading mismatch $|$
6--7: moderate artifacts but plausible $|$
4--5: strong composite feel $|$
1--3: clearly fake.
\end{tcolorbox}

\end{tcolorbox}
\caption{\textbf{LLM evaluation prompt templates (part 1/2).} Both GPT-5.1 and Gemini-3.1-Pro receive the same prompts and return JSON. \texttt{BG\_Cons} compares background regions of the control and generated videos while ignoring the foreground in both. \texttt{FG-BG\_Fusion} judges compositing realism of the generated video alone.}
\label{fig:prompt_llm_part1}
\end{figure*}

\begin{figure*}[!t]
\begin{tcolorbox}[promptbox, title=LLM Evaluation Prompts (Part 2/2): Illum\_Harmony \& Char\_Cons]
\footnotesize

\begin{tcolorbox}[metricbox]
\textbf{Illumination Harmony} (\texttt{Illum\_Harmony}) \hfill \textit{Role: Expert evaluator for lighting consistency.}\\[2pt]
\textit{Input:} A sequence of frames sampled uniformly from one generated video.\\[1pt]
\textit{Task:} Evaluate illumination harmony between the foreground character and the background.\\[1pt]
\textit{Criteria:}
\begin{itemize}[nosep,leftmargin=12pt,label=\textbullet]
\item Lighting direction: key light on subject matches scene cues
\item Color temperature: warm/cool balance matches the background
\item Intensity / exposure: subject brightness matches scene exposure (no ``cutout'' contrast)
\item Shadow plausibility: shadows on the subject (and any cast shadow) fit background lighting
\item Temporal consistency: lighting does not flicker or change unnaturally across frames
\end{itemize}
\textit{Scoring:}
10: perfect lighting match $|$
8--9: minor color-temp / contrast mismatch $|$
6--7: noticeable but not catastrophic $|$
4--5: strong contradiction (direction/temp/exposure clearly off) $|$
1--3: destroys realism.
\end{tcolorbox}

\begin{tcolorbox}[metricbox]
\textbf{Character Consistency} (\texttt{Char\_Cons}) \hfill \textit{Role: Strict evaluator for identity preservation.}\\[2pt]
\textit{Inputs:} (1)~Target reference image(s); (2)~Frames sampled uniformly from the generated video.\\[1pt]
\textit{Task:} Judge how faithfully the generated character matches the target identity. The judge is instructed to be skeptical: passing requires clear, frame-by-frame evidence that the identity is preserved. Sharing only broad casting traits (gender, ethnicity, rough age) is \textbf{not} sufficient.\\[1pt]
\textit{Inspection checklist (across $\geq$3 frames):}
\begin{itemize}[nosep,leftmargin=12pt,label=\textbullet]
\item Face geometry: face shape, jawline, cheekbones, eyes, nose, lips, eyebrows
\item Skin tone, apparent age, hair color / length / style / hairline
\item Facial hair / marks (beard, moles, scars, tattoos)
\item Outfit: every garment's type, color, pattern, fit; accessories (glasses, hats, jewelry)
\item Build / proportions and temporal stability of identity across all frames
\end{itemize}
\textit{Hard caps (apply the lowest triggered):} face geometry differs $\to 4$; skin tone or age clearly differs $\to 3$; outfit clearly differs $\to 5$; both face and outfit differ $\to 3$; identity drifts across frames $\to 4$; obviously a different person $\to 2$; broken/distorted $\to 2$; wrong number of people $\to 2$; original (unswapped) character $\to 1$.\\[1pt]
\textit{Acceptable (do not penalize):} motion blur, mild expression changes, minor texture artifacts, slight color/lighting shifts from scene relighting.\\[1pt]
\textit{Scoring:}
10: indistinguishable from reference $|$
8--9: unambiguously the same person, only subtle deviations $|$
6--7: recognizable but with one clear attribute mismatch $|$
4--5: mixed signals (one major attribute wrong) $|$
2--3: obviously a different person or severely mis-rendered $|$
1: completely different / broken / unswapped.
\end{tcolorbox}

\end{tcolorbox}
\caption{\textbf{LLM evaluation prompt templates (part 2/2).} \texttt{Illum\_Harmony} judges lighting consistency between the rendered character and the scene. \texttt{Char\_Cons} compares the generated character against the target reference image(s) under a strict-evaluation protocol with hard caps on each attribute mismatch.}
\label{fig:prompt_llm_part2}
\end{figure*}

\section{Details on Motion Fidelity Evaluation}
\label{sec:suppl_cotracker}

To evaluate the temporal motion consistency between the generated videos and the ground truth (GT) on the \textbf{Character-Animation Subset}, we employ CoTracker3~\citep{cotracker3} to establish dense, long-term point correspondences. The resulting score is reported as \textbf{Motion Cons.} in Tab.~\ref{tab:quantitative} of the main paper.

\textbf{Foreground-only Tracking.}
Several baselines do not faithfully preserve the original background; comparing trajectories on those regions would unfairly penalize them for an aspect that is unrelated to character motion. To make the motion-fidelity comparison strictly about the character, we therefore restrict the evaluation to the \textbf{foreground human region}. Concretely, we apply the binary character mask of the GT video frame-by-frame to both the GT and the generated videos (background pixels are zeroed), and we further constrain the CoTracker3 query points so that they only lie inside the foreground region of the first frame. This way, all methods are scored only on how well their characters move, regardless of how their backgrounds look.

\textbf{Frame Alignment and Sampling.}
For each video pair, we take the first $80$ frames from both the generated and GT videos, and align them to the same canvas of $1920\times1056$ via cover-resize and center-crop. Before tracking, both videos are then resized to a common compute resolution (short side $384$~px). Query points are sampled inside the foreground mask of the first frame (with a uniform $20\times20$ grid as a fallback when the mask is empty).

\textbf{Joint Visibility Mask.}
CoTracker3 returns trajectory coordinates and per-frame visibility confidences for the GT and the generated videos independently. To prevent occluded or out-of-bounds points from skewing the evaluation, a point's frame is considered valid only if its visibility confidence exceeds $0.5$ in \textit{both} videos simultaneously, and a point is included only if it has at least two such jointly-visible frames.

\textbf{Scale-Invariant Trajectory Comparison.}
A key challenge when evaluating motion fidelity across diverse cinematic scenes is the variation in subject depth, focal length, and absolute motion magnitude. To ensure a fair, normalized comparison across heterogeneous scenes, we evaluate the relative displacement trajectories using \textbf{Trajectory Cosine Similarity}. Let $P_{t}^{(i)}\in\mathbb{R}^{2}$ and $\hat{P}_{t}^{(i)}\in\mathbb{R}^{2}$ be the 2D coordinates of the $i$-th tracked point at frame $t$ for the GT and generated videos, respectively. The displacement relative to the first frame is $\Delta P_{t}^{(i)} = P_{t}^{(i)} - P_{0}^{(i)}$ (and analogously $\Delta\hat{P}_{t}^{(i)}$). Let $V^{(i)}$ be the set of jointly-visible frames for point $i$. We flatten the per-frame 2D displacements over $V^{(i)}$ into single vectors and take their cosine similarity:
\begin{equation}
S^{(i)} = \frac{\sum_{t\in V^{(i)}} \langle \Delta P_{t}^{(i)},\, \Delta\hat{P}_{t}^{(i)} \rangle}
{\sqrt{\sum_{t\in V^{(i)}} \|\Delta P_{t}^{(i)}\|_{2}^{2}}\,\sqrt{\sum_{t\in V^{(i)}} \|\Delta\hat{P}_{t}^{(i)}\|_{2}^{2}}},
\label{eq:cos_sim}
\end{equation}
where $\langle\cdot,\cdot\rangle$ denotes the inner product. The final motion consistency score is the average over all $N_{\mathrm{valid}}$ valid points,
\begin{equation}
\text{Motion Cons.} = \frac{1}{N_{\mathrm{valid}}} \sum_{i=1}^{N_{\mathrm{valid}}} S^{(i)},
\label{eq:motion_score}
\end{equation}
which evaluates the directional and structural fidelity of each motion path while remaining invariant to the absolute pixel distance traveled.

\section{Correlation between Human Preference and Automated Evaluation}
\label{sec:suppl_correlation}

To verify that our LLM-based perceptual evaluation (Sec.~\ref{sec:suppl_llm}) agrees with human consensus, we compute the Spearman rank correlation between the user-study votes and the LLM-based scores at the level of (method $\times$ metric $\times$ \{single, multi\}) cells. For each cell we obtain (i) the human win-rate (the percentage of times each method is voted best by participants) and (ii) the average LLM score from Table~\ref{tab:quantitative} and Table~\ref{tab:gemini}, and pair the two across $14$ method$\times$task tuples (six replacement methods plus eight animation methods). Table~\ref{tab:correlation} reports the per-metric Spearman~$\rho$ separately for the single- and multi-character settings.

\begin{table}[!t]
\centering
\caption{Spearman rank correlation between human preferences (user study) and our LLM-based perceptual scores, across $14$ method$\times$task tuples per metric. Higher values indicate stronger agreement.}
\label{tab:correlation}
\renewcommand{\arraystretch}{1.05}
\setlength{\tabcolsep}{8pt}
\begin{tabular}{lccc}
\toprule
\textbf{Metric} & $n$ & \textbf{Single}~$\rho$ & \textbf{Multi}~$\rho$ \\
\midrule
BG Cons.        & 14 & $+0.407$       & $+0.502$ \\
FG-BG Fusion    & 14 & $\mathbf{+0.859}$ & $\mathbf{+0.856}$ \\
Illum.~Harmony  & 14 & $+0.705$       & $+0.846$ \\
Char.~Cons.     & 14 & $+0.292$       & $+0.581$ \\
\midrule
\textbf{Average} & -- & $\mathbf{+0.566}$ & $\mathbf{+0.696}$ \\
\bottomrule
\end{tabular}
\end{table}

The two strongest agreements are obtained on \textbf{FG-BG Fusion} ($\rho = 0.859$ / $0.856$ for single / multi) and \textbf{Illumination Harmony} ($0.705$ / $0.846$), confirming that the LLM judges share the human notion of compositing realism and lighting plausibility. Background and character-identity consistency, while weaker, still show clear positive correlation with human ratings. Averaged over the four metrics, Spearman~$\rho = 0.566$ on the single-character setting and $\rho = 0.696$ on the multi-character setting, indicating that our automated LLM-based protocol provides a faithful proxy for human perceptual preference.

\begin{figure*}[!tb]
\centering
\animategraphics[width=0.70\linewidth]{5}{fig/supp_swap/supp_swap_}{00}{08}
\caption{Additional qualitative results for the \textbf{Character-Replacement Subset}, showcasing precise identity and background preservation in complex narrative scenes.}
\label{fig:supp_replace}
\end{figure*}

\begin{figure*}[!tb]
\centering
\animategraphics[width=0.80\linewidth]{5}{fig/supp_animation/supp_animation_}{00}{08}
\caption{Additional qualitative results for the \textbf{Character-Animation Subset}, demonstrating high-fidelity motion replication across both single and multi-character scenarios.}
\label{fig:supp_animation}
\end{figure*}

\begin{figure*}[!tb]
\centering
\animategraphics[width=0.75\linewidth]{5}{fig/failure_case/failure_case_}{00}{09}
\caption{Representative failure cases of PAI-Actor. Readers can click and play the video clips in this figure using {\color{red}\textbf{Adobe Acrobat}}.}
\label{fig:failure_case}
\end{figure*}
\section{Computational Cost and Resources}
\label{sec:suppl_cost}

This section reports the additional cost details that are not already covered by Table~\ref{tab:efficiency} and the experimental setup of the main paper.

\textbf{Training Cost.}
Following the experimental setup, all components are finetuned on top of the open-source Wan2.2-TI2V-5B~\citep{wan2025} backbone using LoRA~\citep{hu2022lora} (rank $80$) and AdamW~\citep{loshchilov2019adamw} (learning rate $5\times 10^{-5}$) in BF16 precision, on $8$ NVIDIA H200 GPUs (batch size $1$ per GPU), at native $1920\times 1056$ resolution and $97$ frames. Beyond what the main paper reports, the breakdown across the three training stages is:
\begin{itemize}
    \item Bidirectional in-context teacher: $8{,}000$ steps, ${\sim}215$ hours wall-clock, peak VRAM $\approx$~$136$~GB per GPU.
    \item Stage~1 causal adaptation: $1{,}600$ steps, ${\sim}40$ hours, peak VRAM $\approx$~$92$~GB per GPU.
    \item Stage~2 on-policy self-forcing: $950$ steps, ${\sim}134$ hours, peak VRAM $\approx$~$114$~GB per GPU.
\end{itemize}

\textbf{Inference Cost.}
Inference latency and memory are measured on a single NVIDIA H200 GPU at the resolutions and frame counts reported in Table~\ref{tab:efficiency}. For a $1920\times 1056$, $97$-frame clip:
\begin{itemize}
    \item \textbf{Ours-Bidir} ($50$ steps): $\approx$~$27$~min wall-clock per clip, peak VRAM $\approx$~$136$~GB (FPS $0.06$ in Tab.~\ref{tab:efficiency}).
    \item \textbf{Ours-Causal w/o self-forcing} (causal student, $50$ steps): $\approx$~$16$~min wall-clock per clip, peak VRAM $\approx$~$98$~GB (FPS $0.10$ in Tab.~\ref{tab:efficiency}).
\end{itemize}

For longer videos beyond a single short clip, \textbf{Ours-Causal} extends naturally via KV-cache rolling: a $417$-frame ($\sim$$17$\,s) clip at the same resolution costs $\approx$~$95$~min wall-clock with peak VRAM $\approx$~$100$~GB, whereas \textbf{Ours-Bidir} cannot scale to this length due to the quadratic cost of full bidirectional attention.

The bidirectional teacher prioritizes cinematic-level visual quality and strict temporal coherence within short clips. The autoregressive student, distilled via two-stage self-forcing, retains comparable per-clip fidelity at lower per-clip cost (${\sim}1.7\times$ faster wall-clock and ${\sim}28\%$ less peak VRAM), and uniquely supports KV-cache-based long-video generation that the teacher cannot scale to.

\section{Additional Results}
\label{sec:more_results}

To further demonstrate the versatility and robustness of PAI-Actor, we provide more comprehensive qualitative results. Our evaluation spans four core tasks categorized into two primary applications: \textbf{Character-Replacement Subset} (shown in Fig.~\ref{fig:supp_replace}) and \textbf{Character-Animation Subset} (shown in Fig.~\ref{fig:supp_animation}). Each application includes both \textbf{single-character} and \textbf{multi-character} setups. These results highlight the model's ability to maintain strict identity consistency and temporal coherence, even in highly challenging scenarios featuring complex backgrounds and severe foreground occlusions.

\begin{table}[t]
\centering
\caption{Inference efficiency, tested on one NVIDIA H200.}
\label{tab:efficiency}
\renewcommand{\arraystretch}{1.0}
\footnotesize
\begin{tabular}{lccc}
\toprule
\textbf{Method} & \textbf{Resolution} (W$\times$H) & \textbf{Steps} & \textbf{FPS} $\uparrow$ \\
\midrule
SCAIL                 & $896 \times 512$    & 51 & 0.15 \\
MultiAnimate         & $832 \times 480$    & 50 & 0.13 \\
Wan-Animate           & $1280 \times 720$   & 20 & 0.22 \\
MoCha                 & $832 \times 480$    & 50 & 0.04 \\
VACE                  & $1280 \times 704$   & 50 & 0.18 \\
\midrule
Ours-Bidir            & $1920 \times 1056$  & 50 & 0.06 \\
Ours-Causal w/o self-forcing            & $1920 \times 1056$  & 50 & 0.10 \\
\bottomrule
\end{tabular}
\end{table}

\section{Limitations and Future Work}
\label{sec:limitations}
\label{sec:fail}

While PAI-Actor achieves robust cinematic character replacement, it still faces certain limitations under extreme or highly complex conditions, as illustrated in Figure~\ref{fig:failure_case}.
First, prioritizing high-fidelity generation currently requires extensive denoising steps, leading to higher inference latency compared to real-time systems.
Second, \textbf{scale mismatch} between the reference and target views can cause generation artifacts: if the reference image is a full-body shot but the driving video features a close-up angle, the model may lack sufficient facial prior information, resulting in distorted or blurry faces.
Third, \textbf{severe occlusions and complex multi-character interactions} can cause pose ambiguity. When characters heavily overlap, the extracted skeletal poses may become ambiguous, misguiding the generation process and leading to physically implausible limb rendering.
Fourth, in scenarios with \textbf{high-speed dynamic motions}, fine details may suffer from slight degradation due to the difficulty of maintaining temporal consistency under extreme pixel shifts.
Finally, the \texttt{Ours-Causal w/o self-forcing} faces difficulties with \textbf{out-of-distribution style transfer}, occasionally struggling to faithfully reproduce specific non-photorealistic artistic styles such as 2D anime. Addressing these efficiency and robustness challenges remains an important direction for our future work.

\newpage
\clearpage

\end{document}